\documentclass[lettersize,journal]{IEEEtran}
\usepackage{amsmath,amsfonts}
\usepackage{algorithmic}
\usepackage{algorithm}
\usepackage{array}
\usepackage[caption=false,font=normalsize,labelfont=sf,textfont=sf]{subfig}
\usepackage{textcomp}
\usepackage{stfloats}
\usepackage{url}
\usepackage{verbatim}
\usepackage{graphicx}
\usepackage{cite}
\usepackage{lipsum}
\usepackage{fontawesome} 
\usepackage[dvipsnames]{xcolor}

\NewDocumentCommand{\PawHere}{O{} m}{%
    \begin{flushright}
        {\color{#2} \faPaw\IfValueT{#1}{~#1}}%
    \end{flushright}%
    \vspace{-1mm}%
    {\color{#2}\hrule width \linewidth height 2pt}%
    \vspace{1cm}%
}

\usepackage{color}
\usepackage{soul}
\definecolor{SoftB}{rgb}{0,0.5,1}
\setstcolor{SoftB}

\newif\ifFinalVersion      
\FinalVersionfalse         

\newcommand{\setEqref}[1]{Eq.~(\ref{#1})}

\begin{document}

\title{Latent Mode Decomposition}

\author{Manuel Morante, Naveed ur Rehman,~\IEEEmembership{Senior Member,~IEEE,}
\thanks{The authors are with the Department of Electrical and Computer Engineering, Aarhus University, 8200, Aarhus N, Denmark.}
}

\markboth{Laten Mode Decomposition}%
{\mbox{}}


\maketitle

\begin{abstract}
We introduce Variational Latent Mode Decomposition (VLMD), a new algorithm for extracting oscillatory modes and associated connectivity structures from multivariate signals. VLMD addresses key limitations of existing Multivariate Mode Decomposition (MMD) techniques—including high computational cost, sensitivity to parameter choices, and weak modeling of interchannel dependencies. Its improved performance is driven by a novel underlying model, Latent Mode Decomposition (LMD), which blends sparse coding and mode decomposition to represent multichannel signals as sparse linear combinations of shared latent components composed of AM-FM oscillatory modes. This formulation enables VLMD to operate in a lower-dimensional latent space, enhancing robustness to noise, scalability, and interpretability. The algorithm solves a constrained variational optimization problem that jointly enforces reconstruction fidelity, sparsity, and frequency regularization. Experiments on synthetic and real-world datasets demonstrate that VLMD outperforms state-of-the-art MMD methods in accuracy, efficiency, and the interpretability of extracted structures.
\end{abstract}


\section{Introduction}
\IEEEPARstart{N}{onstationary} signal decomposition techniques constitute an essential tool in Signal Processing for analyzing complex signals. Among them, Mode Decomposition (MD) has emerged as a fundamental framework, enabling the extraction of meaningful intrinsic oscillatory components~\cite{HuaEmp_1998}. Over the last couple of decades, a wide range of MD methods and algorithms have been developed and successfully applied across a wide range of interdisciplinary applications, such as biomedical signal analysis, structural health monitoring, and financial time-series analysis.

This particular trend dates back to the late nineties with the introduction of Empirical Mode Decomposition (EMD) \cite{HuaEmp_1998}, which was followed by the development of other similar alternatives, such as Synchro-squeezed Transform (SST) \cite{DauSyn_2011}, Variational Mode Decomposition (VMD) \cite{DraVar_2014} and Sliding-window Singular Spectrum Analysis (SSA)\cite{HarSli_2018}. Originally designed for single-channel time series analysis, some of these methods were later extended to handle multivariate time series. Notable multivariate algorithms include Multivariate Empirical Mode Decomposition (MEMD)~\cite{RehMul_2009}, multivariate nonlinear chirp mode decomposition~\cite{CheMul_2020}, iterative filtering~\cite{CicMul_2022}, as well as Multivariate Variational Mode Decomposition (MVMD)~\cite{RehMul_2019}.  A more comprehensive comparison of both univariate and multivariate approaches is presented in \cite{EriDat_20222}.

Formally, those Multivariate Mode Decomposition (MMD) approaches share the same fundamental idea; they assume that a multivariate signal of interest accepts a representation as a linear combination of a set of a particular family of amplitude- and frequency-modulated (AM-FM) functions, where all the channels share a well-defined instantaneous frequency \cite{RehMul_2009, RehMul_2019}. In other words, we can decompose a multivariate signal of interest --with $C$ number of channels, $\boldsymbol{x}(t)=[x_{1}(t),x_{2}(t),\ldots,x_{C}(t)]^{\mathsf{T}}$-- into a linear combination of $K$ principal multivariate oscillations as follows:
\begin{equation}
  \label{eq:MMD_model}
  \boldsymbol{x}(t)=\sum_{k=1}^{K}\boldsymbol{u}^{(k)}(t),
\end{equation}
where the $\boldsymbol{u}^{(k)}(t)$, often referred to as the Intrinsic Modes (IMs), constituent modes, or modes, belong to a family of AM-FM functions of the form:
\begin{equation}
  \label{eq:amfm_fun}
  \boldsymbol{u}^{(k)}(t) = \boldsymbol{a}^{(k)}(t)\cos(\phi^{(k)}(t)),
\end{equation}
where $\boldsymbol{a}^{(k)}(t)=[a_{1}(t), a_{2}(t), \ldots, a_{C}(t)]$ and $\phi^{(k)}(t)$ are the instantaneous amplitudes and instantaneous phase associated with the $k$-th mode respectively~\cite{RehMul_2019}. In addition, to ensure these functions represent meaningful oscillations, the amplitudes are always nonnegative, i.e.,  $a_{c}^{(k)}(t)\geq 0\;\forall t$, and the instantaneous frequency, $\omega_{k}(t)=\partial_{t}\phi^{(k)}(t)$, remains positive and exhibits a considerably slower temporal variation compared to the instantaneous phase  $\phi^{(k)}(t)$ \cite{DraVar_2014, RehMul_2019, DauSyn_2011}. 

Intuitively, we can interpret the oscillatory components in Eq.~(\ref{eq:amfm_fun}) as a kind of harmonic-like functions, since they remain concentrated around a central frequency, $\omega_{k}$, while accounting for nonstationary and nonlinear behavior through amplitude and frequency modulations \cite{DraVar_2014, DauSyn_2011}.


Notwithstanding their success, these algorithms face critical limitations in real-world settings. When signals exhibit rich spectral content or moderate nonstationarity, their decomposition accuracy degrades significantly. Moreover, noise contamination, which is common in practical applications, can severely distort the extracted modes, producing misleading or unusable results \cite{Stallone2020}. For example, in MVMD, the decomposition outcome is highly sensitive to user-defined parameters such as the regularization strength and the assumed number of modes. These parameters are typically selected heuristically, which undermines robustness and reproducibility \cite{EriDat_20222}.

Another major limitation is their considerable computational cost. Both MEMD and MVMD operate directly in high-dimensional signal spaces, even though the intrinsic structure of the multivariate signal often lies on a lower-dimensional manifold. In MEMD, mode alignment is achieved through projections along numerous directions (typically 64–128), each requiring a separate univariate EMD computation followed by an averaging step. This redundant processing across projections leads to significant memory and runtime costs \cite{Xun2018}. MVMD, on the other hand, formulates a high-dimensional variational problem over all channels and modes simultaneously \cite{RehMul_2019}. It requires iterative updates of complex-valued analytic signals, mode estimates, and dual variables, resulting in an optimization procedure whose computational cost scales poorly with both the number of channels and the signal length \cite{EriDat_20222}. These factors make existing algorithms impractical for modern high-dimensional multichannel datasets, where scalability and efficiency are essential.

These limitations stem from a deeper issue: existing MMD methods are direct extensions of their univariate counterparts. As a result, they inherit design assumptions that prioritize temporal structure while neglecting the spatial or cross-channel dependencies intrinsic to multivariate signals. Although such methods can align modes across channels based on shared frequency content \cite{RehMul_2009, RehMul_2019}, they lack the modeling capacity to capture the rich inter-channel dynamics present in modern multichannel data, limiting their descriptive and analytical power.

To address the above limitations, we propose a new model for MMD, named Latent Mode Decomposition (LMD), that leverages principles from sparse coding and matrix factorization. Unlike existing MMD approaches, LMD assumes that the observed multichannel signal is not formed independently across channels, but instead arises from a sparse linear combination of a small set of shared \textit{latent components}, each composed of a set of AM-FM \textit{latent modes}. The advantages of the proposed formulation are threefold. First, it provides a principled way to capture interchannel dependencies by explicitly modeling shared latent structures via matrix factorization. Second, representing a signal through a reduced set of latent components improves the model's robustness to noise and enables the model's operation in a lower dimensional space instead of a higher dimensional space, in which MEMD and MVMD operate. Finally, the assumption that the latent components are composed of a linear combination of AM-FM latent modes ensures that the model yields meaningful and interpretable output.


Building on the LMD model, \textit{we propose a simple and efficient algorithm, termed Variational Latent Mode Decomposition (VLMD), to extract latent AM-FM modes and their associated connectivity structures from multivariate data}. To achieve that, we design a variational optimization problem that jointly enforces three key criteria: i) reconstruction fidelity of each observed channel using a shared latent dictionary; ii) sparsity of the mixing coefficients via an $\ell_1$-norm penalty; and iii) constrained frequency behavior of the latent modes, inspired by the bandwidth-regularized formulation in MVMD \cite{RehMul_2019}. To efficiently solve the proposed optimization problem, we derive closed-form update rules for latent components, oscillatory modes, and their associated center frequencies by leveraging the Alternating Direction Method of Multipliers (ADMM). The resulting VLMD algorithm offers several key advantages over state-of-the-art methods, including MVMD and MEMD. 
\begin{itemize}
\item It captures interchannel dependencies more effectively by modeling shared latent structures across channels, and explicitly outputs these connectivity patterns as part of its representation.
\item It exhibits strong robustness to noise, as the sparse coding formulation inherently promotes extracting meaningful components while suppressing irrelevant or noisy content.
\item It delivers substantial computational savings and faster running times, as it operates within the lower dimensional latent representation of the data.
\item It is less sensitive to parameter selection, particularly the number of modes, a critical limitation in MVMD, where this parameter must be manually tuned and is often unknown in practice.
\end{itemize}


The rest of the paper is organized as follows. The theoretical foundations of LMD and its connections to existing decomposition methods are discussed in Section~\ref{sec:LMDmodel}. Section~\ref{sec:ProposedAlgorithm} introduces VLMD, an efficient variational algorithm designed to solve the LMD formulation via sparse coding and constrained frequency modeling. A comprehensive evaluation on synthetic datasets is provided in Section~\ref{sec:SyntheticExperiments}, demonstrating the method’s accuracy, robustness, and computational efficiency. Section~\ref{sec:ExperimentsRealData} showcases the practical utility of VLMD through two real-world applications. Finally, Section~\ref{sec:Conclusions} concludes the paper by summarizing the main contributions and key findings.

\section{Proposed Model and Related Works}
\label{sec:LMDmodel}

\subsection{Latent Mode Decomposition}

Let us assume the observed signal $\mathbf{X} = [\boldsymbol{x}_{1}, \boldsymbol{x}_{2}, \ldots, \boldsymbol{x}_{C}] \in \mathbb{R}^{T \times C}$, which contains $T$ time points across $C$ channels, arises from $K$ sparse linear combinations of $L$ latent modes, with $L\leq C$. In other words, we assume that the matrix $\mathbf{X}$ can be written as:
\begin{equation}
\mathbf{X}=\sum_{k=1}^{K}\mathbf{\Theta}^{(k)}\mathbf{A},
\label{eq:lmd_model}
\end{equation}
where $\mathbf{A} = [\boldsymbol{a}_1, \boldsymbol{a}_2, \ldots, \boldsymbol{a}_C] \in \mathbb{R}^{L \times C}$ is a sparse coefficient matrix, which encodes the connectivity structure between the channels and their latent representations, and $\mathbf{\Theta}^{(k)} = [\boldsymbol{\theta}_{1}^{(k)}, \boldsymbol{\theta}_{2}^{(k)}, \ldots, \boldsymbol{\theta}_{L}^{(k)}] \in \mathbb{R}^{T \times L}$ represents the collection of all the hidden latent modes associated with the $k$-th mode. More specifically, each \(\theta_l^{(k)}\) represents the $k$-th latent mode of the $l$-th latent component, with  
\begin{equation}  
\boldsymbol{\theta}^{(k)}(t) = \boldsymbol{\alpha}^{(k)}(t) \cos(\phi^{(k)}(t)),
\end{equation}  
where $\boldsymbol{\theta}^{(k)}(t) = [\theta_{1}^{(k)}(t), \theta_{2}^{(k)}(t), \ldots, \theta_{L}^{(k)}(t)]^\mathsf{T}$. Moreover, $\boldsymbol{\alpha}^{(k)}(t) = [\alpha_{1}^{(k)}(t), \alpha_{2}^{(k)}(t), \ldots, \alpha_{L}^{(k)}(t)]^\mathsf{T}$, and $\phi^{(k)}(t)$ denote the hidden amplitudes and instantaneous phase of the $k$-th latent mode respectively, similarly as in \setEqref{eq:amfm_fun}.

Alternatively, we can better understand this model by compactly grouping the modes into single latent components (sources), $z_{1}(t), z_{2}(t), \ldots, z_{L}(t)$, where
\begin{equation}  
z_{l}(t) = \sum_{k=1}^{K} \theta_{l}^{(k)}(t), \quad l=1,\ldots,L.
\label{eq:z_def}
\end{equation}
Thus, we can compactly reframe our proposed model from \setEqref{eq:lmd_model} as follows:
\begin{equation}  
  \mathbf{X}=\mathbf{Z}\mathbf{A},
\end{equation}
with $\mathbf{Z} = [\boldsymbol{z}_1, \boldsymbol{z}_2, \ldots, \boldsymbol{z}_L] \in \mathbb{R}^{T \times L}$.

Under this alternative formulation, our proposed model attains a more familiar form, as it aligns with several standard signal processing frameworks~\cite{Sergios_2020}, such as matrix factorization and Dictionary Learning~\cite{MicEla_2010}. Leveraging Dictionary Learning jargon, $\mathbf{Z}$ is referred to as the \textit{dictionary}, where each of its columns contains a specific \emph{atom}~\cite{MicEla_2010}.

Nevertheless, unlike conventional Dictionary Learning approaches, our proposed model assume a very specific constraint over the columns of the dictionary matrix $\mathbf{Z}$. Specifically, \setEqref{eq:z_def} imposes that the latent components, i.e., the atoms of the dictionary, arises from a linear combination of $K$ latent modes. These latent modes fulfill the conditions for a family of low bandwidth AM-FM functions (see Eq. (2) and the related discussion). Further, the spectra of all latent channels associated with the $k$-th mode remain centered around their corresponding central frequency, $w_k$, as originally proposed in \cite{RehMul_2019} for multivariate AM-FM functions.  

\subsection{Conventional MMD: Where are the Intrinsic Modes?} 
\label{sec:LMDandMD} 
One of the key principles of MMD is that the observed signal, say for example at the $c$-th channel, $\boldsymbol{x}_c$, can be decomposed into $K$ different IMs, $\boldsymbol{u}^{(1)}_{c}, \boldsymbol{u}^{(2)}_{c}, \ldots, \boldsymbol{u}^{(K)}_{c}$, see \setEqref{eq:MMD_model}, which constitutes a direct linear approximation of the observed signal. Our proposed LMD preserves the capability of extracting IMs implicitly, even though it does not model them explicitly. In particular, from \setEqref{eq:lmd_model}, one can easily observe that the conventional modes are just encoded by the latent modes and the learned sparse structure within the coefficient matrix, $\mathbf{A}$, as follows:
\begin{equation}
  \label{eq:trans_conventional_modes}
  \mathbf{U}^{(k)} = \mathbf{\Theta}^{(k)} \mathbf{A}\quad k=1,2\ldots,K,
\end{equation}  
where $\mathbf{U}^{(k)} = [\boldsymbol{u}_{1}^{(k)}, \boldsymbol{u}_{2}^{(k)}, \ldots, \boldsymbol{u}_{C}^{(k)}] \in \mathbb{R}^{T \times C}$ contains the IMs associated with the $k$-th frequency across all the $C$ channels.


This is possible as all latent modes associated with the $k$-th mode share the same central frequency, $w_k$. Similarly, this property also highlights the relevance of the coefficient matrix $\mathbf{A}$, which captures channel-specific connectivity structures and reveals how each latent mode contributes to different channels.

\paragraph*{Equivalence between LMD and MMD}  
Interestingly, in the case of independent channels, i.e., when there is no shared structure between channels, we have $\mathbf{A} = \mathbf{I}$. In this case, the latent modes and the IMs become indistinguishable, and our model reduces to conventional MMD.

Nonetheless, this limit case is unrealistic, as most of the real-world multivariate signals of interest typically exhibit some kind of interchannel structure. Thus, LMD appears in this regard as a natural generalization of the MMD principles, as it provides a concrete framework that captures the underlying structure of multivariate signals, while still being able to obtain their IMs from the signal.

\subsection{LMD as frequency-aware Dictionary Learning}  
Until now, we have focused on the MD and frequency-related aspects of LMD, emphasizing its capability to model hidden latent AM-FM components from multivariate signals. However, the proposed LMD model can be interpreted as a specific variant of the classical Dictionary Learning problem~\cite{Sergios_2020, MicEla_2010}. Specifically, in contrast to traditional Dictionary Learning~\cite{MicEla_2010}, we go a step further and propose that the atoms are derived from an additional hidden linear combination of an unknown set of functions, each exhibiting a particular and constrained frequency behavior.

One way to understand our proposed approach is by looking back at the basic ideas behind traditional Fourier-based dictionaries~\cite{CheAto_2001} and wavelet-based Dictionary Learning approaches, where dictionaries are often constructed using wavelet functions to capture multiscale structures in data~\cite{MalWav_2009}. Those methods assume that the signal of interest could be reconstructed using a small set of harmonics (in the case of Fourier dictionaries) or by a set of time-frequency localized wavelets from a predefined mother wavelet~\cite{MalWav_2009}.

In this way, our LMD model shares the same fundamental concepts of capturing essential oscillatory components --with a sufficiently narrow bandwidth-- through a learned sparse representation. However, unlike traditional methods that rely on predefined basis functions, LMD operates fully data-driven, learning the basis functions directly from the data. This allows LMD to adapt to the specific characteristics of the data, resulting in a more flexible and accurate representation of the underlying oscillatory components.

On the other hand, LMD introduces an additional constraint that differs from conventional Dictionary Learning methods: all latent modes associated with the $k$-th frequency exhibit a narrow-banded spectrum centered around the common frequency $w_k$. As a consequence, the atoms become aware of these common central frequencies, giving the dictionary a very specific structure. 

Therefore, this additional imposed internal structure within the dictionary aligns LMD with some modern Dictionary Learning variants. These variants impose their own internal dictionary constraints, such as Convolutional Dictionary Learning~\cite{GarCon_2018} and Deep Dictionary Learning~\cite{TarDee_2016}. While those advanced Dictionary Learning models focus on learning shift-invariant and hierarchical structures within their dictionaries, our LMD model focuses on learning a shared and structured set of narrow-banded oscillatory components.

\section{Proposed algorithm}
\label{sec:ProposedAlgorithm}
\subsection{Variational Latent Mode Decomposition (VLMD)}

We propose a simple and efficient algorithm to extract latent modes through the LMD model given in \setEqref{eq:lmd_model}. For this purpose, we adopt a conventional sparse coding formulation, proposing an optimization task based on the $\ell_1$-norm\footnote{While our current implementation focuses on $\ell_1$-norm regularization, other applications could incorporate alternative sparsity-promoting terms or more efficient optimization strategies. Depending on the particular problem at hand, other alternative well-known sparsity-promoting terms could be also used, such as $\ell_{0}$-norm, or $\ell_{1,0}$-norm \cite{Sergios_2020, MicEla_2010}.} \cite{Sergios_2020, MicEla_2010}. Inspired by the variational solution of MVMD, we implement our solution through a similar variational optimization approach~\cite{RehMul_2019}. We refer to our proposed algorithm as Variational Latent Mode Decomposition (VLMD) in the sequel.

We propose the following optimization formulation for VLMD:
\begin{equation}\label{eq:main_opt}  
  \begin{array}{c}
    {\displaystyle \underset{\{\boldsymbol{z}_{l},\boldsymbol{a}_{c},\boldsymbol{\theta}^{(k)},w_{k}\}}{\text{argmin}}\sum_{c=1}^{C}\left\Vert x_{c}(t)-\sum_{l=1}^{L}a_{lc}z_{l}(t)\right\Vert _{2}^{2}+\lambda\sum_{c=1}^{C}\left\Vert \boldsymbol{a}_{c}\right\Vert _{1}+}\\
    {\displaystyle +\alpha\sum_{l=1}^{L}\sum_{k=1}^{K}\left\Vert \partial_{t}\left[\breve{\theta}_{l}^{(k)}(t)e^{-iw_{k}t}\right]\right\Vert _{2}^{2}\;\text{s.t.}\;|a_{ij}|\leq1\:\forall i,j,}\\
    {\displaystyle \text{and}\;z_{l}(t)=\sum_{k=1}^{K}\theta_{l}^{(k)}(t)}\quad\forall l,
    \end{array}
\end{equation}
where $\breve{\theta}^{(k)}_l$ denotes the Hilbert Transform of $\theta^{(k)}_l$, and $\partial_{t}$ stands for the derivative with respect to time. 

The first two terms correspond to the approximation of the signal in its sparse representation, which employs the classical $\ell_1$-norm sparsity-promoting term.

The next term captures the oscillatory behavior of the latent modes, which adopts a similar form as the constraint employed in MVMD~\cite{RehMul_2019}. Intuitively, this term ensures the latent modes remain relatively close to some common central frequency, $w_k$. This constraint works together with the dictionary constraint, ensuring that the signal representation emerges from a linear combination of the latent modes.

Finally, note that we have included an additional constraint on the coefficient matrix, $|a_{ij}|\leq 1$, which prevents degenerate solutions and enhances interpretability, as matrix factorization yields equivalent solutions under different scaling factors~\cite{Sergios_2020, MicEla_2010}.

\subsection{Solution of the optimization task}

Solving the optimization task in \setEqref{eq:main_opt} poses significant challenges due to the inherent complexity of the problem and the constraints involved. However, the literature offers several strategies to tackle such optimization tasks effectively. In particular, following the classical divide-and-conquer approach, we employ the Alternating Direction Method of Multipliers (ADMM)~\cite{BoyDis_2011}, a well-established method for solving constrained optimization problems. Additionally, this choice aligns with various approaches in Dictionary Learning, e.g.,~\cite{GiaSim_2016}, where ADMM has been successfully applied to efficiently integrate multiple constraints.

Following the ADMM approach~\cite{BoyDis_2011}, we first construct the augmented Lagrangian of the problem. Thus, we introduce the dual variable $\boldsymbol{\gamma}(t) = [\gamma_1(t), \gamma_2(t), \ldots, \gamma_L(t)]^\mathsf{T}$ and the penalty parameter $\rho > 0$. After making the necessary algebraic manipulations on \setEqref{eq:main_opt}, we obtain the following augmented Lagrangian: 


\begin{align*}
  L_{\rho}&=\sum_{c=1}^{C}\left\Vert x_{c}(t)-\sum_{l=1}^{L}a_{cl}z_{l}(t)\right\Vert _{2}^{2} + \lambda \sum_{c=1}^{C} \left\Vert \boldsymbol{a}_{c}\right\Vert _{1}+\\
  & + \alpha\sum_{k=1}^{K}\sum_{l=1}^{L}\left\Vert \partial_{t}\left[\breve{\theta}_{l}^{(k)}(t)e^{-iw_{k}t}\right]\right\Vert _{2}^{2} -\frac{\rho}{2}\sum_{l=1}^{L}\left\Vert \gamma_{l}(t)\right\Vert _{2}^{2}+ \\
  & +\frac{\rho}{2}\sum_{l=1}^{L}\left\Vert z_{l}(t)-\sum_{k=1}^{K}\theta_{l}^{(k)}(t)+\gamma_{l}(t)\right\Vert _{2}^{2}.
\end{align*}

Next, we follow the ADMM framework to decompose the main optimization problem into a sequence of the following iterative suboptimization tasks:

\subsubsection{Optimization with respect to the coefficients $\boldsymbol{a}_{c}$}
First, we optimize with respect to the coefficients for each channel by solving the following optimization task:
\begin{equation}
  \boldsymbol{a}_{c}^{\star}=\underset{\boldsymbol{a}_{c}}{\text{argmin}}\quad L_{\rho}(\{\boldsymbol{a}\}) \quad\text{s.t.}\quad |a_i|\leq 1 \;\forall i.
\end{equation}

By using matrix notation, and after some algebraic manipulations, this problem can be compactly written as the matrix factorization problem:
\begin{equation}
  \label{eq:SparseUpdate}
  \mathbf{A}^{\star}=\underset{\mathbf{A}}{\text{argmin}}\left\Vert \mathbf{X}-\mathbf{Z}\mathbf{A}\right\Vert _{F}^{2}+\lambda\left\Vert \mathbf{A}\right\Vert _{1}\,\text{s.t.}\,|a_{ij}|\leq1\;\forall ij.
\end{equation}

Through fixing the variables $\mathbf{X}$ and $\mathbf{Z}$, this problem constitutes a well-established sparse decomposition problem ~\cite{Sergios_2020, TibReg_1996}. Further, several algorithms and efficient implementations exist to solve this optimization task~\cite{MeiRel_2007, MicEla_2010, Sergios_2020}.

\subsubsection{Optimization with respect to the latent components $z_{l}(t)$}
For this stage, considering the $l$-th latent component, we have the following optimization task:
\begin{equation}
z_{l}^{\star}(t)=\underset{z_{l}}{\text{argmin}}\quad L_{\rho}(\{z_{l}(t)\}).
\end{equation}

By excluding all terms that do not contribute to the $l$-th latent signal, we obtain:  
\begin{align*}
\underset{z_{l}(t)}{\text{argmin }}\sum_{c=1}^{C}\left\Vert x_{c}(t)-\sum_{n=1}^{L}a_{nc}z_{n}(t)\right\Vert _{2}^{2}+\\
+\frac{\rho}{2}\left\Vert z_{l}(t)-\sum_{k=1}^{K}\theta_{l}^{(k)}(t)+\gamma_{l}(t)\right\Vert _{2}^{2}.
\end{align*}

For convenience, let us define the residual function, $r_{cl}$, and the auxiliary function, $h_{l}$, as follows:
\begin{align} 
  r_{cl}(t)&=x_{c}(t)-\sum_{n\neq l}^{L}a_{nc}z_{nc}(t),\\
    h_{l}(t)&=\sum_{k=1}^{K}\theta_{l}^{(k)}(t)-\gamma_{l}(t).
\end{align}

Using this notation, the optimization task can be compactly written as:
\begin{equation}
  \underset{z_{l}(t)}{\text{argmin}}\sum_{c=1}^{C}\left\Vert r_{cl}(t)-a_{lc}z_{l}(t)\right\Vert +\frac{\rho}{2}\left\Vert z_{l}(t)-h_{l}(t)\right\Vert _{2}^{2}.
\end{equation}

While we could attempt a direct solution on this stage, we found that optimizing in the Fourier domain is more efficient, leading to a more computationally efficient algorithm. Thus, applying the Plancherel Theorem, we reformulate the optimization task equivalently in the Fourier domain as follows:
\begin{equation}
  \label{eq:LatentSignalFourier}
  \underset{\hat{z}_{l}(\omega)}{\text{argmin}}\sum_{c=1}^{C}\left\Vert \hat{r}_{c}(\omega)-a_{lc}\hat{z}_{l}(\omega)\right\Vert _{2}^{2}+\frac{\rho}{2}\left\Vert \hat{z}_{l}(\omega)-\hat{h}_{l}(\omega)\right\Vert _{2}^{2},
\end{equation}
where we introduced the notation $\hat{f}$ to denote the Fourier transform of $f$.

Formally, \setEqref{eq:LatentSignalFourier} constitutes a variational problem over the Fourier domain, which closely resembles the variational problem studied by \cite{DraVar_2014}. Interestingly, we can solve this problem directly following similar steps, yielding the following closed-form solution:
\begin{equation}
  \label{eq:LatentSignal}
  \hat{z}_{l}^{\star}(w)=\frac{\frac{2}{\rho}\sum_{c=1}^{C}a_{lc}\hat{r}_{c}(w)+\sum_{k=1}^{K}\hat{\theta}_{l}^{(k)}(w)-\hat{\gamma}_{l}(w)}{1+\frac{2}{\rho}\sum_{c=1}^{C}a_{lc}^{2}}.
\end{equation}

\subsubsection{Optimization with respect the latent modes $\boldsymbol{\theta}^{(k)}$}

In this stage, we focus on updating the latent modes $\boldsymbol{\theta}^{(k)}$, with $k = 1, 2, \ldots, K$, across all latent channels. For the $k$-th frequency within the $l$-th latent channel, the optimization problem takes the form:  
\begin{equation}
  \theta_{l}^{\star (k)} = \underset{\theta_{l}^{(k)}}{\text{argmin}} \quad L_{\rho}(\{\boldsymbol{\theta}\}).
\end{equation}

By excluding all terms that do not contribute to the $l$-th latent mode associated with the $k$-th frequency, we obtain:
\begin{align*}
  \underset{\theta_{l}^{(k)}}{\text{argmin}}\ \alpha\left\Vert \partial_{t}\left[\breve{\theta}_{l}^{(k)}(t)e^{-iw_{k}t}\right]\right\Vert _{2}^{2}+ \\ 
  +\frac{\rho}{2}\left\Vert z_{l}(t)-\sum_{q=1}^{K}\theta_{l}^{(q)}(t)+\gamma_{n}(t)\right\Vert _{2}^{2}.
\end{align*} 

The resulting subproblem resembles the one addressed by \cite{RehMul_2019}, but here we seek the representation of the latent signal rather than the original signal. Yet, despite this minor difference, our subproblem effectively renders the same optimization task as (16) in \cite{RehMul_2019}. Therefore, following the same steps as prescribed by \cite{RehMul_2019}, we can obtain the following closed-form solution:
\begin{equation}
  \label{eq:LatentModes}
  \hat{\theta}_{l}^{(k)\star}(\omega) = \frac{  
\hat{z}_{l}(\omega) - \sum_{q \neq k} \hat{\theta}_{l}^{(q)}(\omega) + \hat{\gamma}_{l}(\omega)  
}{  
1 + \frac{4 \alpha}{\rho} (\omega - \omega_{k})^{2}  
} 
\end{equation}

Observe that unlike the MVMD based formulation in \cite{Reh_Mul_2019}, our formulation simultaneously incorporates both regularization parameters, $\alpha$ and $\rho$, as jointly optimizing the latent representation and mode decomposition.

\subsubsection{Optimization with respect the central frequencies}
At this step, we focus on minimizing the objective with respect to the frequency of the $k$-th channel, $w_k$, while keeping the remaining variables fixed. We can express the corresponding optimization problem as follows:
\begin{equation}
  w_k^\star =\underset{w_k}{\text{argmin}} \quad L_{\rho}(\{\boldsymbol{w}\})
\end{equation}
In this case, only a subset of the terms in the objective function depends on the variable $w_k$. This observation simplifies the optimization task to:
\begin{equation}
  w_k^\star=\underset{w_k}{\text{argmin}}\;\sum_{l=1}^{L}\left\Vert \partial_{t} \left[ \breve{\theta}_{n}^{(k)}(t) e^{-i w_{k} t} \right] \right\Vert _{2}^{2}
\end{equation}
This optimization problem can be directly sovled over the spectral domain, as illustrated in (30)-(32) in \cite{RehMul_2019}, which leads to the solution:
\begin{equation}
  w_k^\star=\frac{\sum_{l}^{L}\intop_{0}^{\infty}w|\hat{\theta}_{l}^{(k)}(w)|^{2}dw}{\sum_{l}^{L}\intop_{0}^{\infty}|\hat{\theta}_{l}^{(k)}(w)|^{2}dw}
\end{equation}

\subsubsection{Update stage for the dual variables}
Finally, for the introduced dual variable, we have the update stage

\begin{equation}
  \boldsymbol{\gamma}^{\star}_{l}(t)=\boldsymbol{\gamma}_{l}(t)+\boldsymbol{z}_{l}(t)-\sum_{k=1}^{K}\boldsymbol{\theta}^{(k)}_{l}(t)\quad \forall l,
\end{equation}
which constitutes the running sum of the residuals.

\subsection{Pseudocode and implementation}
We have developed an efficient implementation of our algorithm in Python. The complete source code is publicly available on GitHub\footnote{VLMD: \url{https://github.com/Dmocrito/vlmd}\label{fn:VLMD_repository}}. The pseudocode outlining the main steps of the algorithm is presented in Algorithm \ref{alg:algVLMD}.

\algsetup{
  linenosize=\small,
  linenodelimiter=.
}
\begin{algorithm}[t]
\caption{Variational Latent Mode Decomposition}\label{alg:algVLMD}
\begin{algorithmic}[1]
\REQUIRE Data matrix $\mathbf{X}\in\mathbb{R}^{T\times C}$, $L$ and $K$
\STATE \textit{Parameters:} $\alpha$, $\rho$, $\lambda$
\STATE \textit{Stopping threshold:} $tol$
\STATE \textbf{Initialize}
\STATE $\mathbf{A}_{ij} \gets  \delta_{ij}\quad \forall i,j$
\STATE $\boldsymbol{z}_l \gets \boldsymbol{x}_l$, $\hat{\boldsymbol{\theta}}_{l}^{(k)} \gets \boldsymbol{0}$, $w_{k}\gets 0\quad \forall l,k$
\STATE $\tau \gets 0.9$
\REPEAT
  \STATE \textit{Sparse coding stage Eq.~(\ref{eq:SparseUpdate})}
  \STATE $\mathbf{A} \gets$ LASSO$_{\lambda}(\mathbf{X},\mathbf{Z})$
  \FOR{$c = 1:C$}
    \STATE $\boldsymbol{a}_c \gets c$-th column of $\mathbf{A}$ 
    \STATE $\boldsymbol{a}_c \gets \frac{\boldsymbol{a}_{c}}{\text{max}\{1, \text{max}(\boldsymbol{a}_{c})\}}$
  \ENDFOR
  \STATE \textit{Latent signal update}
  \FOR{$l = 1:L$}
    \STATE $\hat{\boldsymbol{z}}_{l} \gets \boldsymbol{0}$
    \STATE $\hat{\mathbf{R}}\gets\hat{\mathbf{X}}-\hat{\mathbf{Z}}\mathbf{A}$
    \STATE $\boldsymbol{a}_{l}\gets$ $l$-th row of $\mathbf{A}$
    \STATE $\hat{\boldsymbol{z}}_{l}\gets\frac{\frac{2}{\rho}\boldsymbol{a}_{l}^{\mathsf{T}}\hat{\mathbf{R}}+\sum_{k}\boldsymbol{\theta}_{l}^{(k)}-\hat{\boldsymbol{\gamma}}_{l}}{1+\frac{2}{\rho}\boldsymbol{a}_{l}^{\mathsf{T}}\boldsymbol{a}_{l}}$
  \ENDFOR
  \STATE \textit{Latent modes update}
  \FOR{$k = 1:K$}
    \FOR{$l = 1:L$}
      \STATE $\hat{\boldsymbol{\theta}}_{l}^{(k)}\gets\boldsymbol{0}$
      \STATE $\hat{\boldsymbol{\theta}}_{l}^{(k)}\gets\frac{\hat{\boldsymbol{z}}_{l}-\sum_{i}\hat{\boldsymbol{\theta}}_{l}^{(i)}+\hat{\boldsymbol{\gamma}}_{l}}{1+4\frac{\alpha}{\rho}(w-w_{k})^{2}}$
    \ENDFOR
    \STATE $\eta_{k} \gets \omega_{k}$ set auxiliary variable
    \STATE \textit{Update central frequencies}
    \STATE $w_{k}\gets\frac{\sum_{l}\left(\sum_{w}w|\hat{\theta}_{l}^{(k)}(w)|^2\right)}{\sum_{l}\sum_{w}|\hat{\theta}_{l}^{(k)}(w)|^{2}}$
  \ENDFOR
  \STATE \textit{Update dual variables}
  \FOR{$l = 1:L$}
    \STATE $\hat{\boldsymbol{\gamma}}_{l}\gets\hat{\boldsymbol{\gamma}}_{l}+\tau\left(\hat{\boldsymbol{z}}_{l}-\sum_{k=1}^{K}\hat{\boldsymbol{\theta}}_{l}^{(k)}\right)$
  \ENDFOR
  \STATE \textit{Update convergence diff}
  \STATE $dif \gets \sum_{k} |\eta_{k}-\omega_{k}|^2$
\UNTIL{$dif \leq tol$} 
\end{algorithmic}
\label{alg:LMD}
\end{algorithm}


\subsection*{Implementation Notes}
For the sparse coding stage (line 9 in Algorithm \ref{alg:algVLMD}), we employed the LASSO algorithm from \texttt{scikit-learn}\footnote{\url{https://scikit-learn.org/stable/modules/generated/sklearn.linear_model.Lasso.html}}. This implementation is known for its robustness and scalability, and it delivers strong performance even on large datasets, rendering an excellent candidate for this step. 

To enforce the additional constraint over the coefficient matrix, we exploited the equivalence under scaling factor of the matrix factorization problem to ensure that the coefficients remained within the defined range (see line 11). Similarly, this rescaling fits smoothly within the ADMM framework.

In addition to the described approach for the sparse coding step, we explored other alternative optimization libraries. In particular, we tested the library \texttt{CVXPY}\footnote{\url{https://www.cvxpy.org/}}, which constitutes a popular tool for solving certain constrained convex problems.

For small examples with relatively low number of channels, we observed that both implementation rendered to similar performance. However, for larger datasets, the \texttt{CVXPY} implementation failed to scale efficiently and was considerably slower than the LASSO algorithm. Therefore, we decided to discard it in favor of the LASSO algorithm.


Regarding the update of the latent modes, although it bears some resemblance to the approach in \cite{RehMul_2019}, we opted for a different strategy: drawing inspiration again from \cite{RubEff_2008}, we implement a similar approach, which significantly reduces the overall computational cost.

Furthermore, several loops involved in updating the latent modes and dual variables --specifically those in lines 22 and 32 in Algorithm \ref{alg:algVLMD}-- where efficiently implemented using vectorized operations. Those loops are explicitly stated in Algorithm \ref{alg:algVLMD} to facilitate understand the implementation. Overall, this vectorized operations further enhances both the scalability and runtime performance of the algorithm, particularly in high-dimensional settings.

Finally, we introduced an additional positive step-size parameter, $\tau\in [0,1]$, to regulate the convergence speed of the algorithm, effectively acting as a learning rate parameter. Although this parameter has no influence on the final solution, it acts as a tuning knob to control the rate of convergence, which may be useful in some scenarios.

\section{Experimental results with synthetic data}
\label{sec:SyntheticExperiments}
In this section, we present a detailed analysis comparing the performance of the proposed VLMD algorithm with two state-of-the-art methods: MVMD\footnote{MVMD: We used the open Python implementation: \url{https://github.com/Dmocrito/mvmd}  \label{fn:MVMD_repository}} and MEMD\footnote{MEMD: We used the open Python implementation: \url{https://github.com/mariogrune/MEMD-Python-} \label{fn:MEMD_repository}}. This analysis was conducted using synthetic datasets, allowing for controlled and reproducible conditions under which the effectiveness of each method was systematically evaluated.

\subsection{Synthetic datasets}
To evaluate the performance of the examined methods, we generated several synthetic datasets using a Python script based on the LMD model. This script enables the generation of signals with diverse characteristics, including varying noise levels, frequency distributions, and different levels of amplitude and frequency modulation. The synthetic data generator is openly available on the VLMD's repository\footref{fn:VLMD_repository}.


For simplicity and clarity, we focused on study three representative scenarios. Table~\ref{tab:SynthParameters} summarizes the key parameters for each case that we used for generating those datasets.

\begin{table}[!t]
  \caption{Parameters selected for the three representative scenarios.\label{tab:SynthParameters}}
  \centering
  \begin{tabular}{lccc}
     &  \textbf{Scenario A} & \textbf{Scenario B} &  \textbf{Scenario C}\tabularnewline
   \hline 
   \hline 
   Channels & 5 & 5 & 100\tabularnewline
   Latents & 3 & 3 & 35\tabularnewline
   Sparsity & 0.6 & 0.6 & 0.6\tabularnewline
   Modes ($K$) & 5 & 5 & 5\tabularnewline
   Freq. (Hz) & 5, 17, 50, 73, 110 & 7, 12, 61, 73, 79 & 7, 12, 61, 73, 79\tabularnewline
   AM & (2, 2) & (2, 2) & (2, 2)\tabularnewline
   FM & No & (1, 3) & (1, 3)\tabularnewline
   Noise
    & 0.01 - 10 & 0.01 - 10 & 0.01 - 10\tabularnewline
   \hline 
  \end{tabular}
\end{table}

First, scenario A represents a simple case, involving a few quasi-harmonic functions and channels. The central frequencies are also evenly distributed across the spectrum of the signal. Scenario B depicts a more realistic setting, where the signal consists of a mixture of AM-FM components. In addition, some frequencies appear closely spaced\footnote{This consideration on distinct frequency configurations is critical, as MVMD and MEMD have shown difficulties on unveiling components when some central frequencies appear close to each other.}, reflecting a common condition in many practical scenarios. Finally, scenario C builds upon scenario B by introducing a larger number of channels. 

For each scenario, we generated 10 different datasets using several combinations of amplitude- and frequency-modulations and varying their mix coefficients to minimize the impact of any peculiar configuration. In addition, we repeated each experiment five times with different random noise seeds, across various noise levels, to ensure that stochastic noise did not introduce bias or distort the outcomes.

\subsection{General performance evaluation}
In each studied scenario (see Table~\ref{tab:SynthParameters}), we assessed the performance of the proposed VLMD and compared it against MVMD\footref{fn:MVMD_repository} and MEMD\footref{fn:MEMD_repository}. Specifically, we evaluated two key aspects: (1) the correlation between the extracted IMs, which reflects the methods' ability to isolate relevant signal modes among channels, and (2) the frequency Mean Absolute Percentage Error (MAPE), which quantifies how accurately each method recovers the true frequencies. Finally, recall that VLMD extracts the IMs indirectly through the transformation in \setEqref{eq:trans_conventional_modes}.

\begin{figure*}[!t]
  \centering
  \includegraphics[width=1\textwidth]{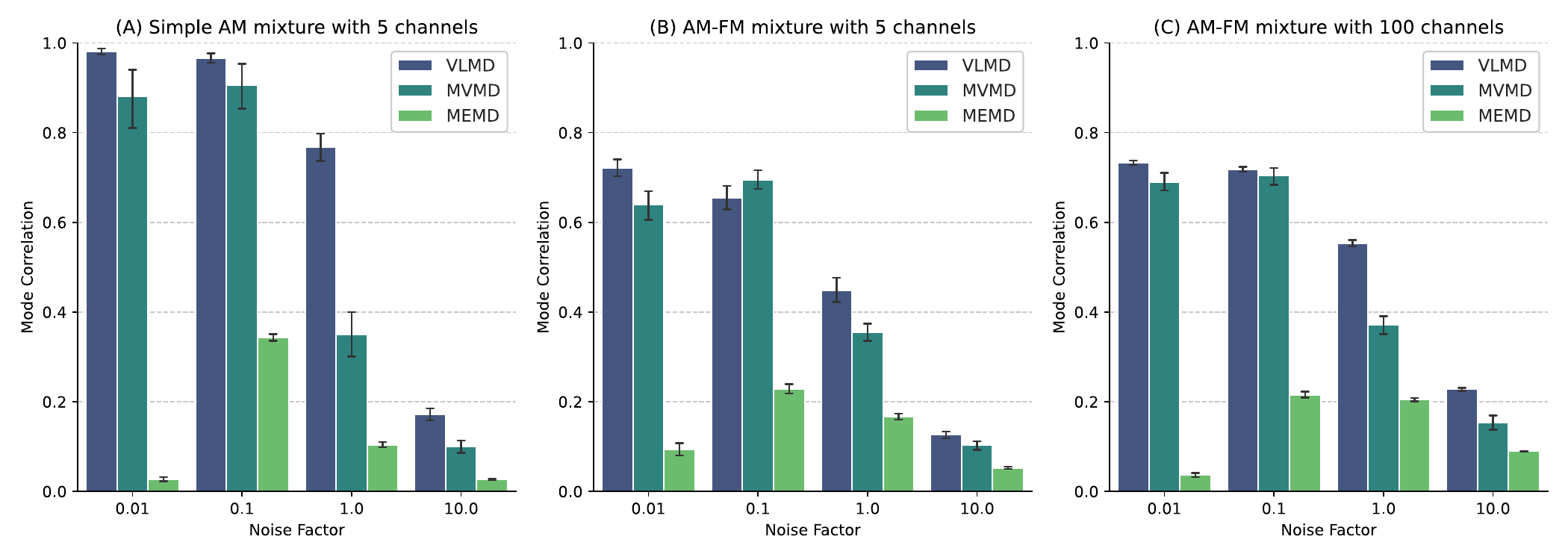}
  \vspace{-3mm}
  \includegraphics[width=1\textwidth]{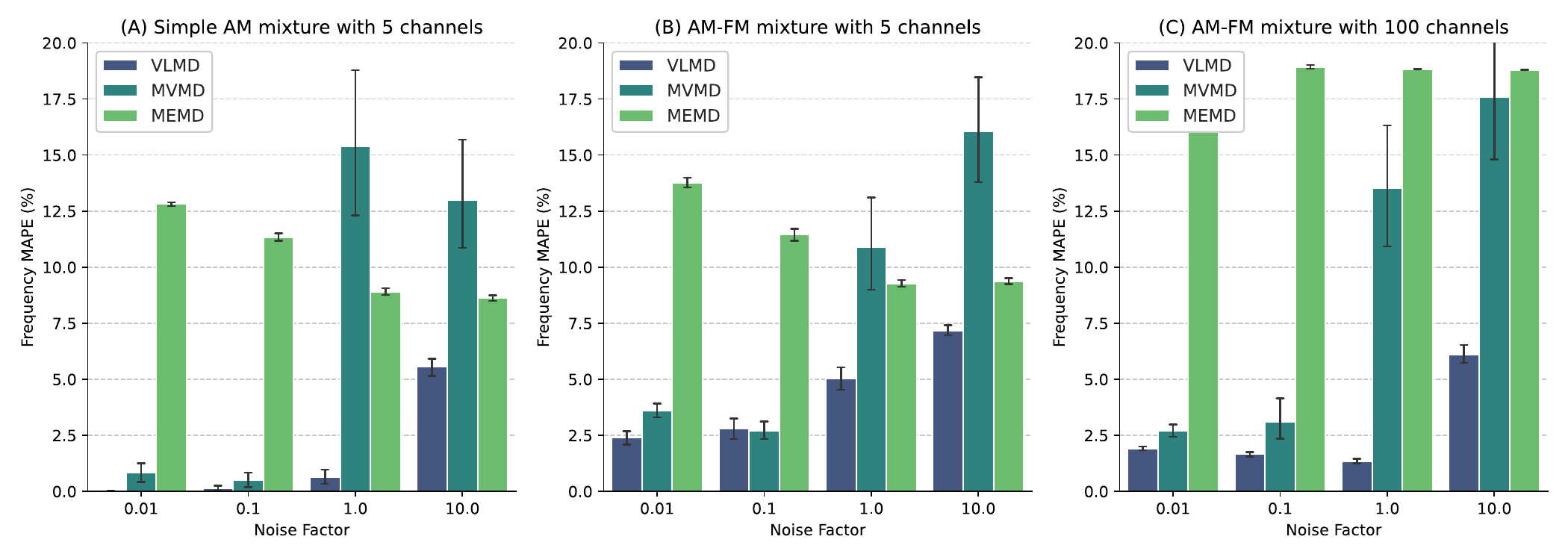}
  \caption{\textbf{Performance comparison.} IMs correlation error and frequency MAPE for the MEMD, MVMD and VLMD algorithms across the three main studied scenarios for different level of noise. Each bar shows the average results corresponding to the 10 different studied cases for each studied scenario and noise seeds.}
  \label{fig:ErorrComparison}
\end{figure*}

Figure~\ref{fig:ErorrComparison} summarizes the correlation error and the frequency MAPE across all the studied scenarios at different levels of noise. For each studied case, we selected the optimal parameters that rendered the optimal results for each algorithm. This process was possible in this case because we had access to the synthetic ground truth. 

In general, in cases with minimal noise, both VLMD and MVMD showed comparable performances. Particularly, in scenario A, MVMD and VLMD both provided excellent results. However, we observed some relevant differences: surprisingly, even in the simplest scenario with low noise VLMD performed better than the MVMD, even though VLMD obtains the IMs indirectly. Similarly, as the noise level increased, MVMD's performance deteriorated rapidly, while VLMD still provided reliable results. This degradation was more severe when we looked at frequency errors, where the sharp deterioration of MVMD compared to VLMD appeared more pronounced. 

Regarding MEMD performance, we observed that this algorithm considerably struggled with unveiling the modes of interest and retrieving the central frequencies, even in the simplest experiment, as illustrated in  Figure~\ref{fig:ErorrComparison}. One of the reasons behind this poor performance is that we are far from a simple combination of sinusoids~\cite{RehMul_2009}. Instead, we targeted a more realistic scenario using AM and AM-FM mixtures.

Finally, when considering the overall performance of all the studied scenarios, we observed that VLMD exhibited robust behavior under varying noise conditions. Even in the challenging noisy scenarios, where MVMD and MEMD completely fell apart, VLMD consistently achieved the highest correlation and --most importantly-- a considerably lower frequency MAPE. These results suggest that VLMD appears a promising tool for signal processing tasks, especially in environments with significant noise.

\subsection{Frequency evolution and convergence} 
To further understand the behavior of VLMD and MVMD, we shifted our focus to examining the frequency evolution and convergence characteristics of the algorithms under study. This exploration aims to provide deeper insights into how each algorithm adapts and unveils different central frequencies per iteration under varying noise levels. MEMD was excluded from this comparison because its performance is incompatible with this particular analysis, as MEMD extracts one mode at a time in a greedy fashion~\cite{RehMul_2009}.

\begin{figure}[!t]
  \centering
  \textbf{Scenario A}
  \includegraphics[width=\columnwidth]{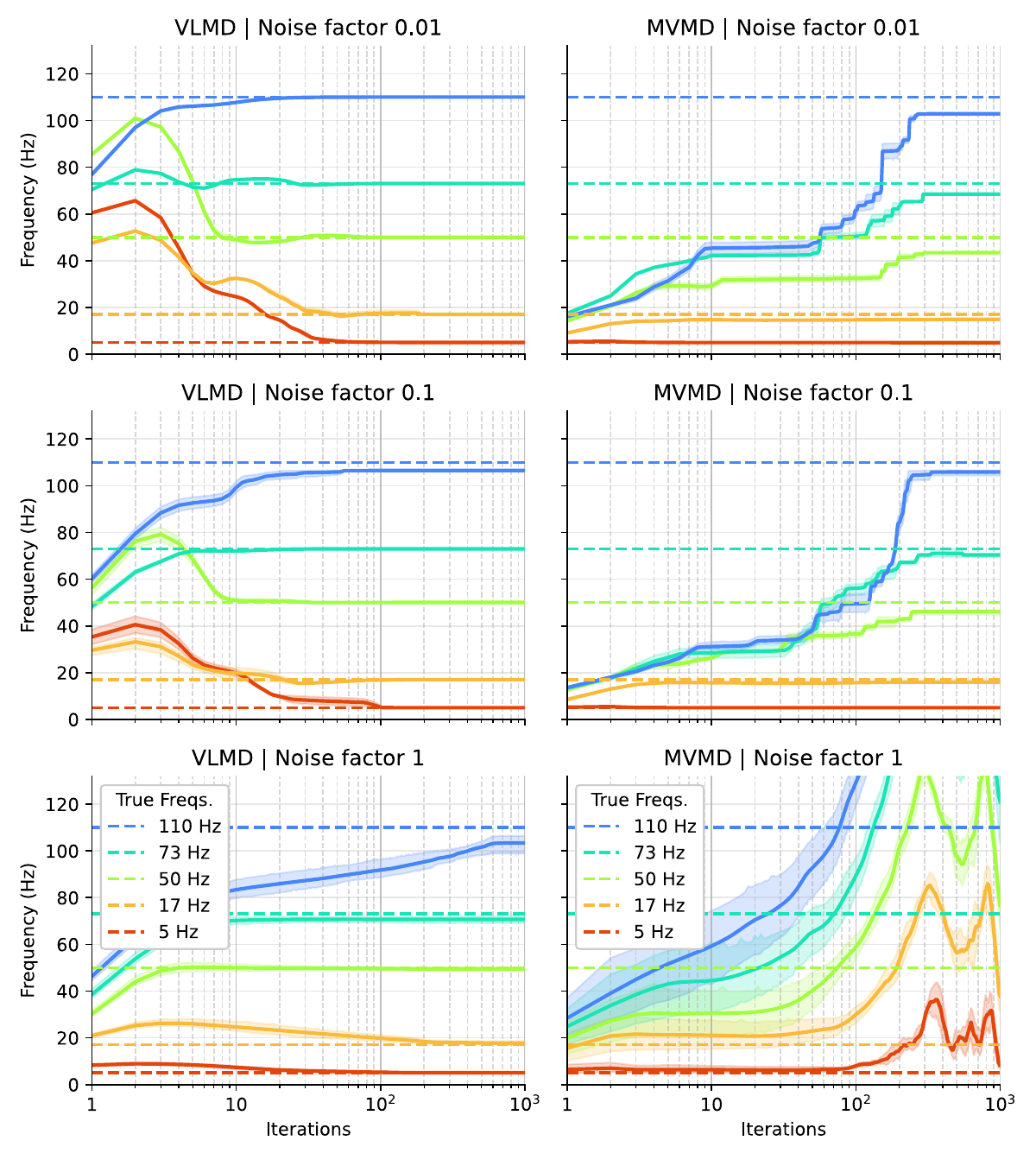}
  \textbf{Scenario B}
  \includegraphics[width=\columnwidth]{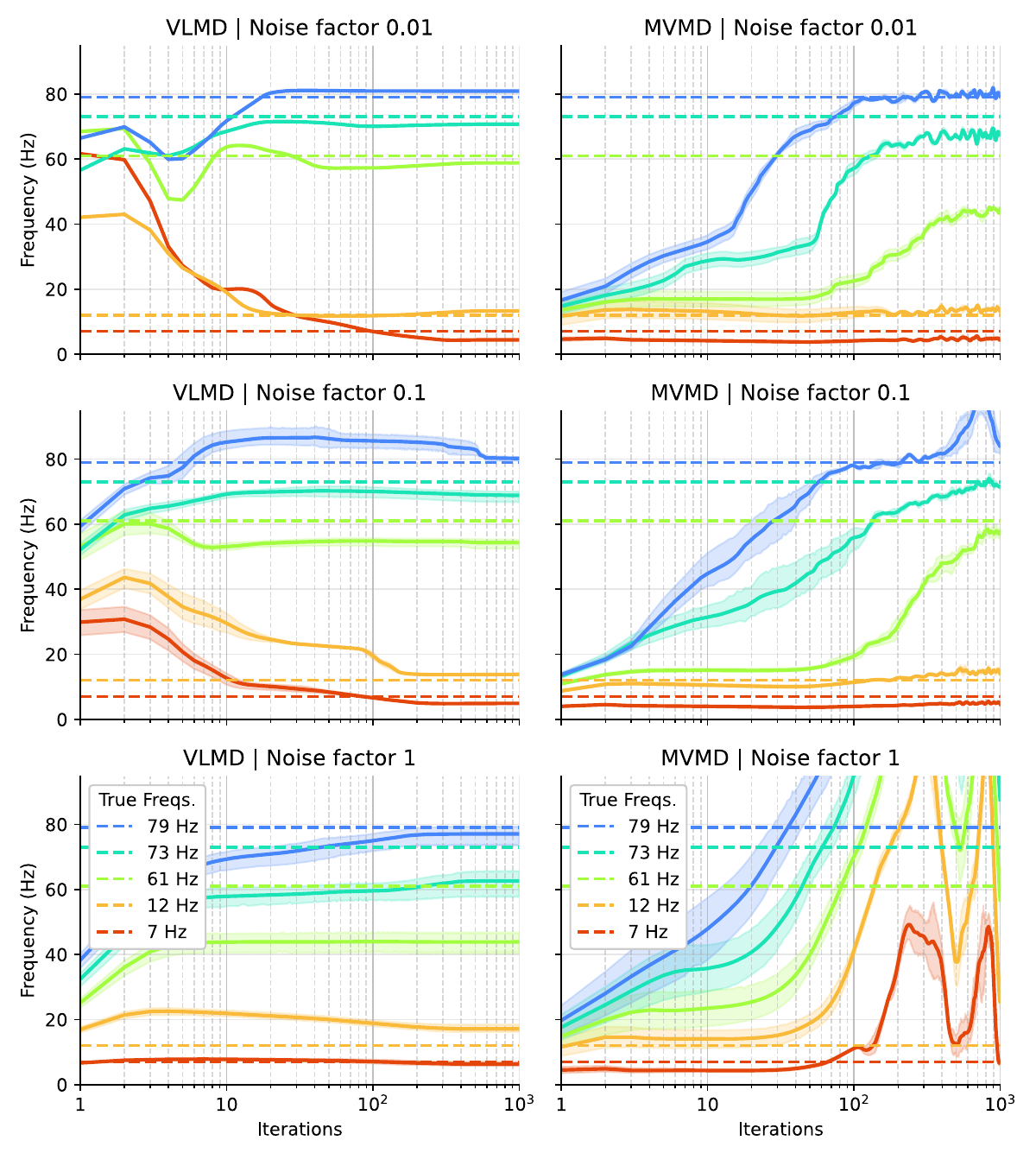}
  \caption{\textbf{Estimated frequencies per iteration.} Evolution of the central frequencies, $\omega_{k}$, estimated by VLMD and MVMD respectively per iteration for Scenario A and B for different level of noise. Each line depicts the average central frequency estimated by each algorithm per iteration for all the studied cases. The horizontal dashed lines correspond to the ground truth.}
  \label{fig:FreqEvolution}
\end{figure}

Figure~\ref{fig:FreqEvolution} illustrates the progression of central frequency estimations per iteration of VLMD and MVMD. For simplicity, we depicted the results corresponding to scenarios A and B for different levels of noise. Each colored line represents the central frequency associated with each model, while the horizontal dashed color lines correspond to the ground truth used to generate the synthetic data.

In general, Figure~\ref{fig:FreqEvolution} exposed that the studied algorithm produced distinct convergence behavior. Firstly, when looking at the results from scenario A, both algorithms seem to effectively converge at the correct central frequencies. Nonetheless, we can see that MVMD requires considerably more number of interations than VLMD. Similarly, in the case of high levels of noise, we can see that MVMD fails to uncover the correct frequencies.

In addition, the computational improvement is not only reflected on the number of iterations. On average, we observed that VLMD on average was almost twice as fast as MVMD. This difference was even more noticeable when working with scenario C, where the number of modes was considerably lower than the number of channels, showing the significant computational advantages of working within the lower-dimensional space.

On the other hand, the results from scenario B in Figure~\ref{fig:FreqEvolution} showed the effects of having an AM-FM with a realistic frequency distribution: it shows how the MVMD struggled to find the appropriate central frequency when those appeared relatively close to each other. We also observed that the problem considerably worsened as noise increased: at higher levels of noise, the MVMD's frequency estimates became unstable, indicating a complete lack of convergence and rendering the MVMD entirely noise-driven. VLMD, on the other hand, remained stable and accurate, converging to values near the ground truth even at severe noise levels, i.e., even under high-noise conditions, it still provides meaningful and stable estimates. 

Overall, these results collectively demonstrate the effectiveness and robustness of the proposed VLMD algorithm. Particularly, these findings suggest that VLMD may be more suitable for real-world applications where data is often noisy and complex. Its stability and accuracy under challenging conditions make it a reliable choice.

\subsection{Parameter sensitivity evaluation}
We investigated the sensitivity of the VLMD and MVMD algorithms to selection of the number of modes, $K$. We focused on this parameter, as it is the most relevant and critical in practical applications. For simplicity, we only studied the data from scenario A, as correspond to the most simple scenario. For the evaluation, we compared the results obtained from each algorithm against the ground truth when the number of modes was overestimated.

Regarding the error calculations, we conducted an optimal search to identify optimal matching between the calculated IMs and the ground truth using the Hungarian algorithm~\cite{KuhHun_1955}. This ensured that, despite having more estimates than ground truth IMs, each estimated IM was compared to its closest ground truth counterpart. The remaining unmatched ones were considered noise residuals. For all the studied cases, we fine-tuned the rest of the parameters that produced the most accurate results for both studied algorithms, compared to the ground truth.

\begin{figure*}[!t]
  \centering
  \includegraphics[width=\textwidth]{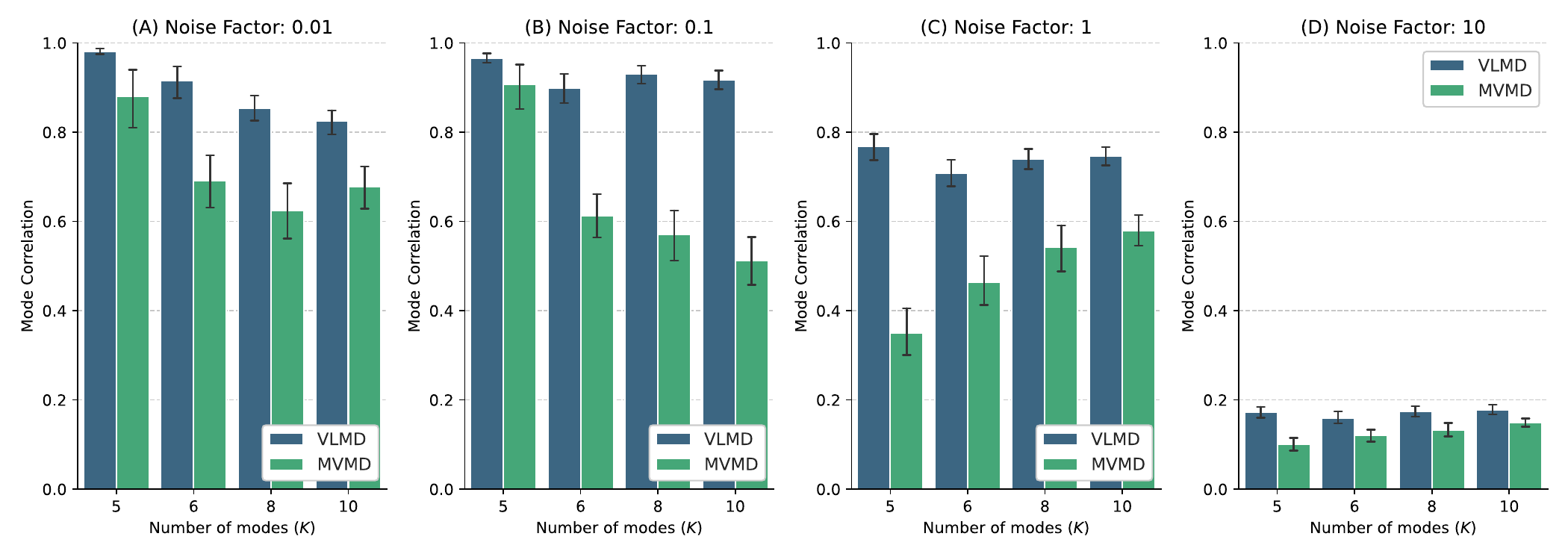}
  \caption{\textbf{Overestimation of the number of modes.} IMs correlation error of VLMD and MVMD algorithms for varying numbers of selected modes. The results belong to the analysis of the data from Scenario A, where $K=5$ corresponds to the ground truth (see Table~\ref{tab:SynthParameters}). The figure also displays several panels (A-D), corresponding to varying noise levels. Each bar represents the average correlation error for 5 different studied cases with different initialization seeds.}
  \label{fig:OverK}
\end{figure*}

Figure~\ref{fig:OverK} illustrates the correlation error for VLMD and MVMD for different selected numbers of modes. Each subplot depicts the effects across several noise levels. In general, we observed a significant degradation in the MVMD's performance as the number of modes was overestimated, i.e., $K>5$, even at relatively low noise levels. Likewise, we found that the MVMD was extremely sensitive to the particular selection of this parameter: even with the overestimation of a single additional mode ($K=6$), the MVMD's performance dropped significantly, regardless of noise level.

In contrast, VLMD demonstrated stable and consistent behavior even when the number of modes $K$ was overestimated. These results highlight the robustness of VLMD in contrast to MVMD's fragile behavior. This robustness of VLMD suggests its superior reliability in real-world applications where the number of modes is unknown.

Another noteworthy result appeared in Figure~\ref{fig:OverK}-B, where we observed that the outcomes in this case were better than the results in the low-noise case in Figure~\ref{fig:OverK}-A. This evidences that the presence of a small amount of noise improved performance when the number of modes was overestimated. In other words, the proposed approach appeared to absorb the extra overestimated component as noise, while preserving the integrity of the remaining components.

The observed improvement in the presence of small noise levels is not unusual. A number of dictionary applications have documented the described situation in practical applications~\cite{MicEla_2010}; noise and an overestimation of components are often beneficial~\cite{Sergios_2020} as they allow noise components to be taken into account during the decomposition process.

\section{Experimental study with real data}
\label{sec:ExperimentsRealData}

We further examined the performance of the VLMD algorithm using two real-world data studies. Through these examples, we provided insight into the applicability, interpretability and performance of VLMD beyond controlled synthetic experiments.

\subsection{Exchange rate dataset}
For the first real-data study, we used the popular exchange rate dataset\footnote{\url{https://github.com/laiguokun/multivariate-time-series-data?tab=readme-ov-file}}, which contains the collection on the daily exchange rates of eight foreign countries including Australia, British, Canada, Switzerland, China, Japan, New Zealand and Singapore ranging from 1990 to 2016. This dataset contains the daily exchange rates fluctuations over a 26-year period, offering a long-term overview of the evolution of their financial markets and economic interdependencies.

\subsubsection*{Data analysis} For this study, we focused primarily on identifying and interpreting the main fluctuations across the studied countries. To ensure a consistent comparison, we first removed the mean value from the data for each country, centering the time series around zero before applying VLMD. This step is important, as it eliminates baseline differences and emphasizes relative variations. 

Regarding the selection of parameters  --for this study-- we selected five latent channels. This ensured that the decomposition had enough latent channels to capture the underlying structure. Additionally, we selected eight different modes, as this value provided sufficient interpretable modes. We also experimented with larger values, both for the number of latent channels and modes, yet these resulted in similar results. 

For the remaining VLMD's hyperparameters, after some empirical testing, we adopted the configuration $\alpha=1000$, $\rho=0.6$, and $\lambda = 0.04$. This selection yielded a stable and interpretable decomposition.

\subsubsection*{Key findings} 
\begin{figure}[!t]
  \centering
  \includegraphics[width=\columnwidth]{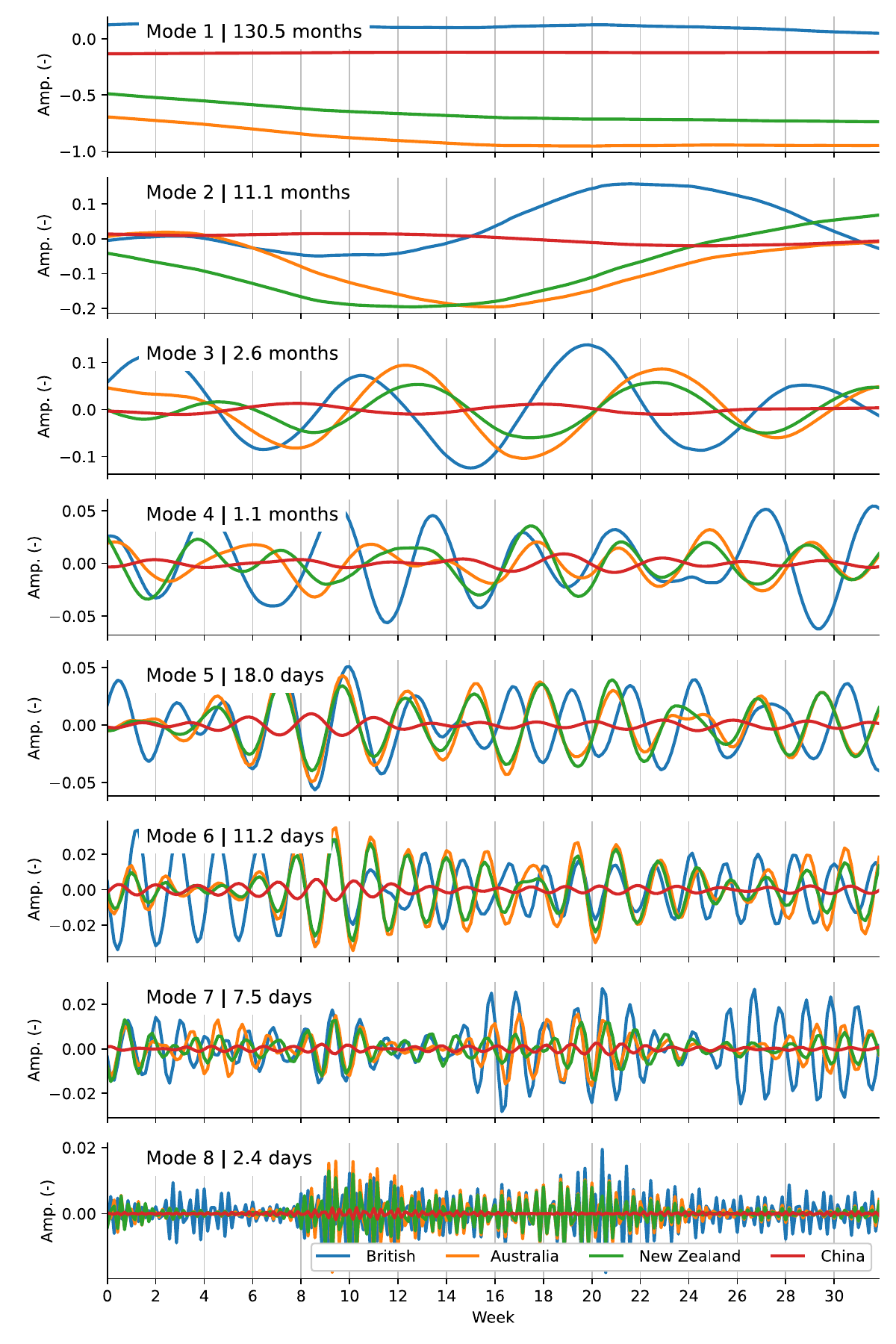}
  \caption{\textbf{IMs for four selected countries.} This figure illustrates the obtained IMs for four selected countries. Each subplot also depict the main period (central frequency) associated with each IM.}
  \label{fig:EcoModes}
\end{figure}

Figure \ref{fig:EcoModes} illustrates the obtained IMs for four selected countries over a randomly chosen 30-week window. The figure also indicates the main periods (central frequency) associated with each IM. Overall, the extracted periods correspond to standard financial business cycles. For instance, IMs 2, 3, 4 and 7 exhibit periods of approximately one year, three months, one month and one week, which align well with typical financial reporting and economic activity cycles. More specifically, IM 3 revealed quarterly patterns consistent with standard financial reporting intervals, while IM 7 showed a clearly weekly pattern. Thus, this suggests that the extracted IMs effectively captured natural economic activity across the studied countries.

\begin{figure}[!t]
  \centering
  \includegraphics[width=0.7\columnwidth]{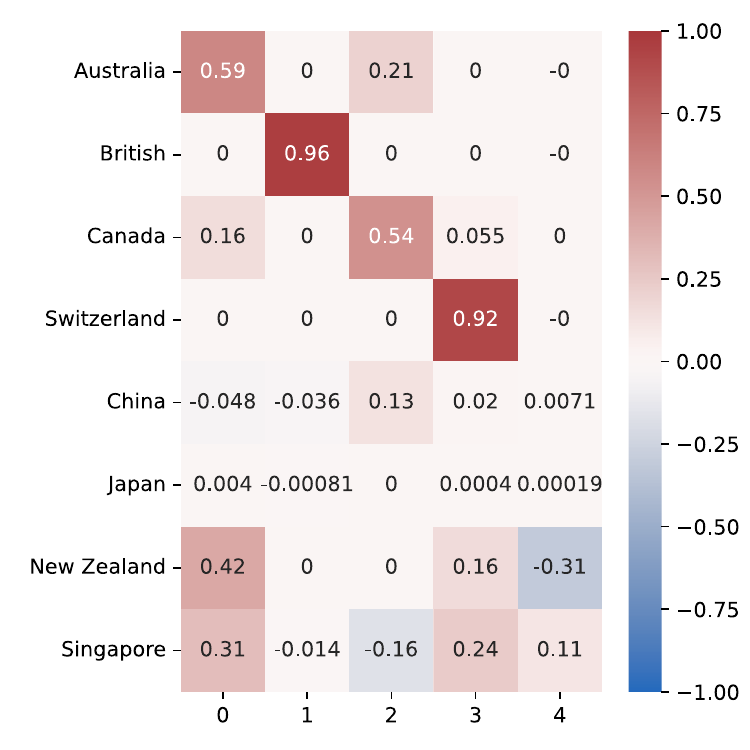}
  \caption{\textbf{Coefficient matrix of the exchange rate dataset.} Heatmap of the values from the coefficient matrix from VLMD. Each row corresponds to a specific channel from the data associated with each studied country, whereas each column corresponds to a specific latent channel.}
  \label{fig:EcoCoefs}
\end{figure}

On the other hand, unlike other popular MMD algorithms, one of the advantages of the proposed approach is that VLMD captures internal dependencies among channels. In this case, VLMD allowed us to investigate how countries' exchange rates are interconnected and their potential relationships. To this aim, we analyzed the obtained coefficient matrix. Figure~\ref{fig:EcoCoefs} depicts the actual values of the coefficient matrix, which works as a feature map, unveiling the connections and interactions between countries and their latent channels. For example, Britain and Switzerland appear largely influenced by a single, nearly isolated latent component. This indicates that their exchange rate is minimally influenced by the rest of studied countries' economic changes in the dataset. On the other hand, Australia, New Zealand and Canada all share several common latent components, indicating a more interconnected economic relationship.

\begin{figure}[!t]
  \centering
  \includegraphics[width=0.9\columnwidth]{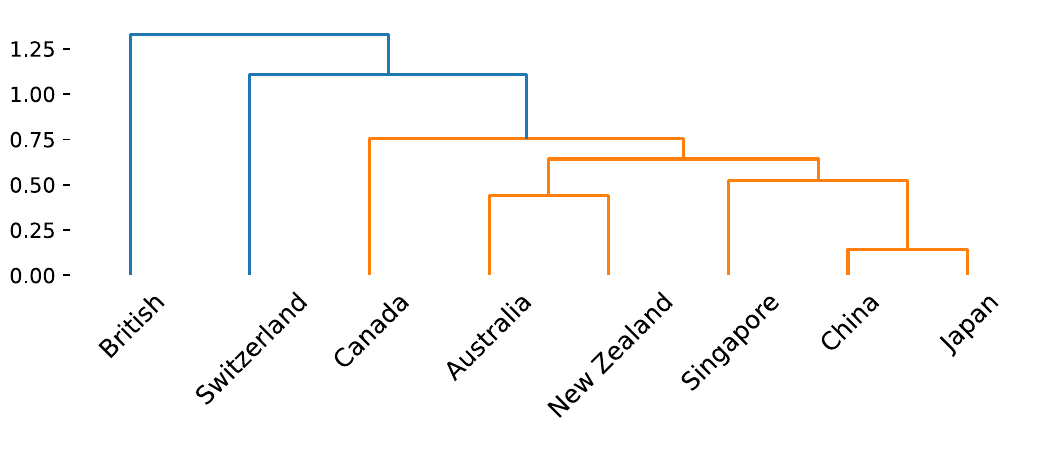}
  \caption{\textbf{Cluster analysis of the coefficient matrix.} Using the coefficient matrix as a feature map, this figure depicts a cluster analysis among rows of the coefficient matrix, using the Euclidean distance.}
  \label{fig:Eco-dendrogram}
\end{figure}

To further examine these qualitative observations, we conducted a cluster analysis using the Euclidean distance between rows of the coefficient matrix to assess similarities, where the latent channels were considered as distinct features. Figure~\ref{fig:Eco-dendrogram} shows the results of this cluster analysis, which confirms the initial qualitative observations from Figure~\ref{fig:EcoCoefs}. Additionally, this analysis revealed that China and Japan also exhibit a closely related behavior, followed by Singapore, possibly due to their unique monetary and regional financial integrations.

In summary, the analysis of the IMs highlights the cyclical nature of exchange rates, while the coefficient matrix provides insights into the interconnectedness and interdependencies between countries. The coexistence of these two dimensions highlights the multiscale and multivariate nature of the exchange rate, emphasizing the heterogeneity and interconnected structure of global markets.

\subsection{Electric grid consumption}

In this second study, we used the openly available Electricity Load Diagram 2011-2014 dataset\footnote{\url{https://archive.ics.uci.edu/dataset/321/electricityloaddiagrams20112014}}, which contains electricity load measurements from 370 clients from Portugal over a four-year period. The dataset contains a diverse range of clients, from households to industrial consumers. It offers a high-resolution temporal profile suitable for oscillatory analysis, such as mode decomposition.

\subsubsection*{Data analysis} In this study, we examined the general consumption patterns and behavior of consumption. Therefore, we excluded any clients with highly irregular or minimal energy consumption Specifically, we first discarded the first year and the last 6 months of the dataset, as some clients were not yet connected to the grid. Additionally, we excluded all clients with a no-consumption period longer than 1\%. These criteria led to a refined dataset of 251 clients. Finally, as we are only interested in the fluctuations of the consumption, we normalized the clients' data to have zero mean and unitary standard deviation. 

Without any further preprocessing, we input the selected clients' data into the VLMD algorithm. For this analysis, we observed that 30 latent channels with 10 modes, with $\alpha =5000$, $\rho=5$, and $\lambda = 0.5$. While we explored other alternative settings, we found that comparable configurations led to similar results, we found that the selected configuration lead to the most interpretable results.

\begin{figure}[!t]
  \centering
  \includegraphics[width=\columnwidth]{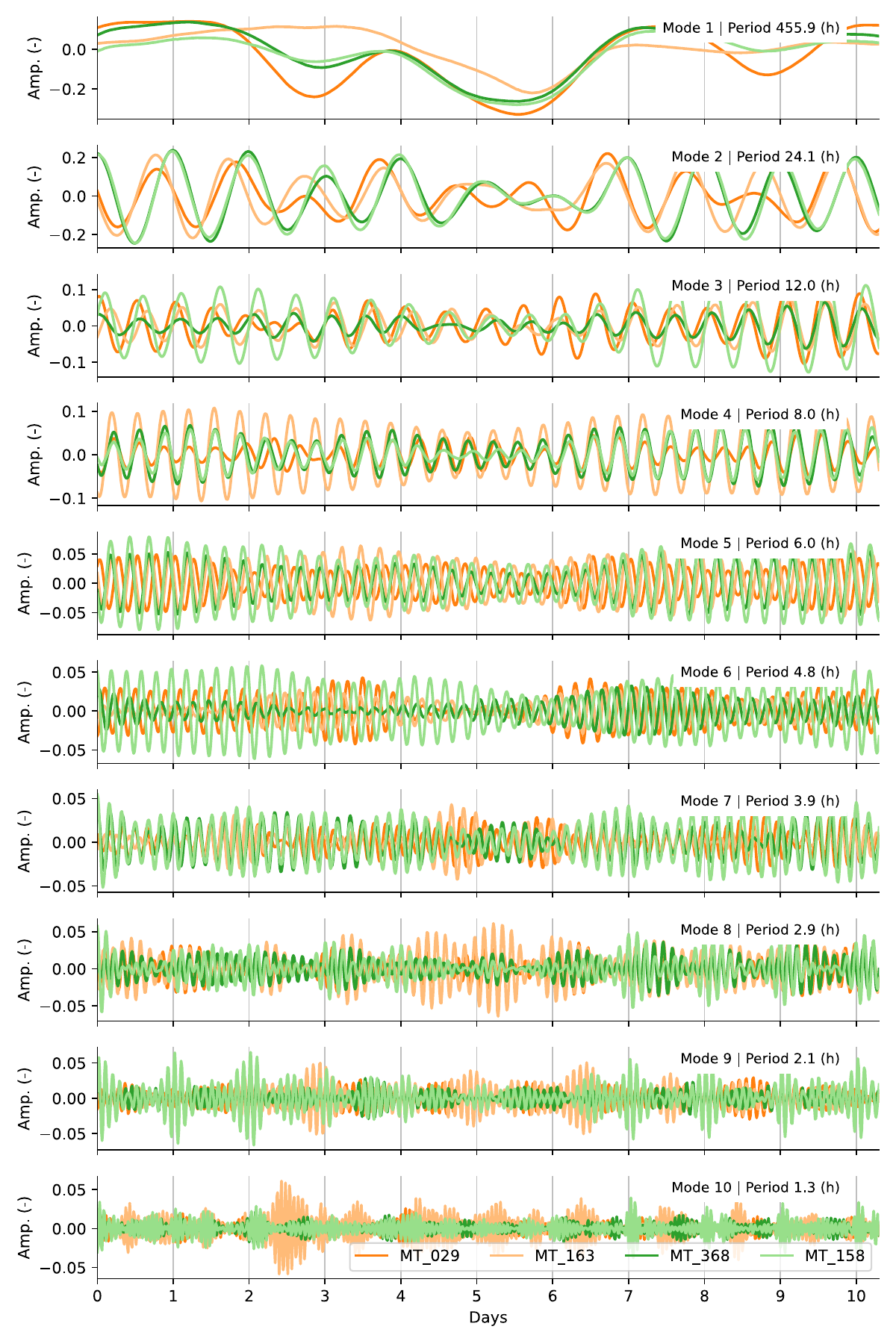}
  \caption{\textbf{Electricity consumption IMs from four selected clients.} This figure shows the obtained IMs after applying the VLMD algorithm to the electricity consumption data. Each colored line corresponds to a selected client from the 251 clients in the studied dataset. The graph shows the IMs over a randomly chosen period of 10 days.}
  \label{fig:ElectricGrid}
\end{figure}

\subsubsection*{Key findings} Figure~\ref{fig:ElectricGrid} shows the obtained IMs with their corresponding period in hours\footnote{We decide to show the period associated with each mode instead of their corresponding frequency because it facilitates the interpretation of the results for this particular dataset.} for a randomly selected 10-day window. Each colored line represents the consumption associated with each conventional mode for a small selection of clients.

\begin{figure}[!t]
  \centering
  \includegraphics[width=\columnwidth]{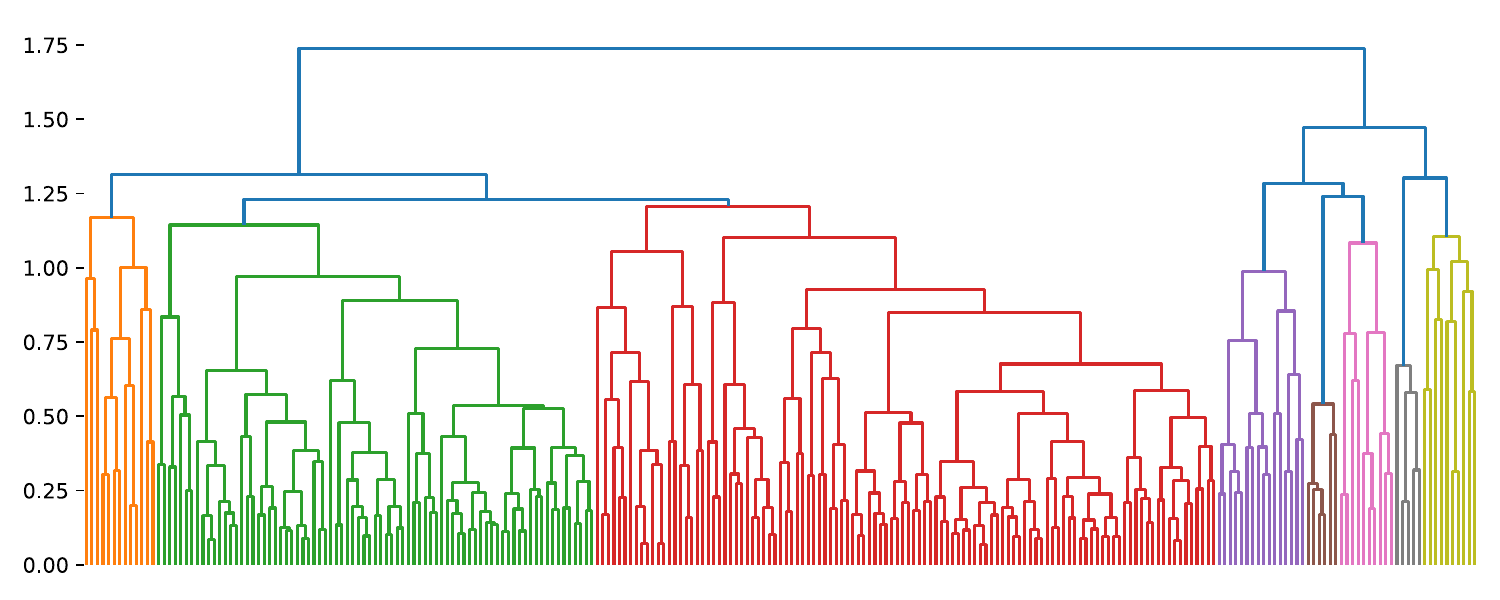}
  \caption{\textbf{Cluster analysis for the 24-h period mode.} This figure shows the results of the clustering analysis based on the correlation between the 24h IMs across all the clients. For simplicity, the figure only depicts the main clusters up to the level 50 of the dendrogram, as very similar clients were grouped together. }
  \label{fig:elgrid_dendo_coef}
\end{figure}

The most remarkable result was the emergence of an IM with a consistent 24h periodicity. This daily periodicity was expected, as daily cycles are a defining feature of electricity usage patterns tied to routine human and industrial activity patterns. The second and third most significant components correspond to a 12h and 6h cycle. Those cycles are also natural as they may reflect additional subdaily routines, such as activity peaks or shifts in industrial operations. Beyond these, other smaller periodic components were also observed, albeit with lower amplitudes as shown in Figure~\ref{fig:ElectricGrid}.

\begin{figure}[!t]
  \centering
  \includegraphics[width=\columnwidth]{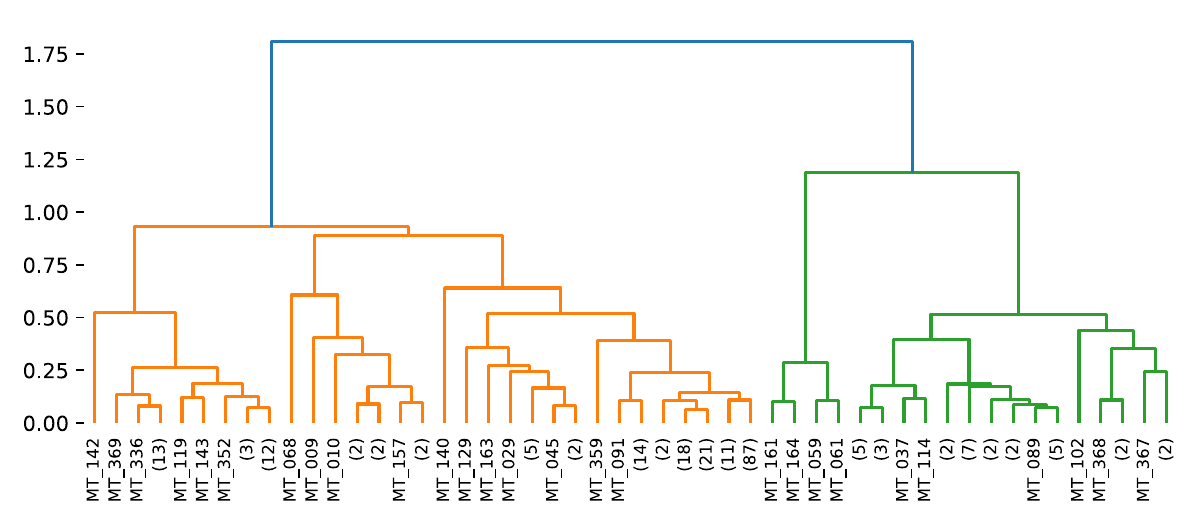}
  \caption{\textbf{Cluster analysis for the 24-h period mode.} This figure shows the clustering of the obtained modes based on the correlation between the IMs associated with the 24h period. For simplicity, this figure only depicts 50 levels of the dendrogram.} 
  \label{fig:elgrid_dendrogram}
\end{figure}

Following the same procedure as with the previous studied example, we can analyze the connections and relationships between clients by examining the obtained coefficient matrix from VLMD. To this aim, we performed also a clustering analysis of all clients. Figure~\ref{fig:elgrid_dendo_coef} depicts the results of this analysis. The results indicate that some groups of clients within the dataset exhibit similar patterns of electricity consumption.

To further investigate the 24 h cycle, we conducted a simple clustering analysis based on the correlation between the extracted modes. Figure~\ref{fig:elgrid_dendrogram} shows a dendrogram based on the correlation between the modes for the 24 h period. These results pointed out that the clients appeared to be segregated into two primary groups, with distinct consumption behaviors. 

A closer inspection of the results in Figure~\ref{fig:ElectricGrid}, shows that the green lines (MT 158 and MT 368) correspond to clients within the green cluster, whereas the orange lines (MT 029 and MT 163) belong to the orange cluster. When focusing on the 24h IMs, we observed that the main difference between the clusters is that the IMs are shifted by approximately 4 hours. Meanwhile, clients within the same cluster exhibit high correlation with one another, indicating synchronized consumption patterns. This temporal offset suggests that while both clusters follow similar daily cycles, their peak usage times vary, potentially due to differing operational schedules or sector-specific demands. We suggest that these two clusters may correspond to distinct usage contexts, specifically, residential and work-related consumption patterns.


\section{Conclusions}
\label{sec:Conclusions}
We have introduced the Latent Mode Decomposition (LMD) model, which elegantly blends the principles of sparse coding and matrix factorization with multivariate mode decomposition to effectively extract relevant information from structured multivariate signals. Building upon the basis of the proposed model, we have also proposed a novel algorithm --Variational Latent Mode Decomposition (VLMD)-- for efficiently decomposing multivariate signals into their inherent latent modes and components. 

Extensive experiments demonstrated VLMD's robustness to noise compared to MVMD and MEMD, two of the most popular MMD alternatives. In addition, VLMD showed reduced sensitivity to parameter selection, especially regarding the number of intrinsic modes (IMs). On the other hand, MVMD exhibited high sensitivity to mode selection. This was observed even in cases with reduced noise levels, which highlights the advantages of VLMD in real-world applications where the exact number of modes is unknown.

Two study cases using real-world data further illustrate VLMD's potential as a reliable alternative to analyzing multivariate signals. In the first example, the proposed algorithm effectively captured the natural oscillatory component associated with expected standard financial cycles. At the same time, it allowed us to further explore the interconnections between economies. This revealed that some economies display synchronous fluctuations, while others operate more independently. The coexistence of these two factors highlights the exchange rate's multiscale and multivariate nature. In the second example, VLMD successfully extracted meaningful patterns from electric grid consumption. The obtained results provided natural cycles, and their corresponding modes allowed us to easily classify different customers into meaningful clusters.

Overall, the demonstrated success of VLMD in these applications suggests a promising path for future research across diverse domains, such as system monitoring, financial data analysis, electric grid studies or biomedical signal processing. Future work may also extend the framework to handle other types of structured signals or support more complex representations by incorporating alternative sparsity-promoting constraints.

\bibliographystyle{IEEEtran}
\bibliography{IEEEabrv,references}

 




\vfill

\end{document}